\documentclass[pdflatex,sn-mathphys]{sn-jnl}
\usepackage{csquotes}

\jyear{2021}%

\theoremstyle{thmstyleone}%
\theoremstyle{thmstyletwo}%

\theoremstyle{thmstylethree}%

\newcounter{notecounter}

\newcommand{\enoteson}{\long\gdef\enote##1##2{{
			\stepcounter{notecounter}
			\large\bf
			\hspace{100cm}\arabic{notecounter} $<<<$ ##1: ##2
			$>>>$\hspace{1cm}}}}
\enoteson

\begin{document}

\title[Addressing the Challenges of Cross-Lingual Hate Speech Detection]{Addressing the Challenges of Cross-Lingual Hate Speech Detection}

\author*[1]{\fnm{Irina} \sur{Bigoulaeva}}\email{www.ukp.tu-darmstadt.de}

\author[2]{\fnm{Viktor} \sur{Hangya}}\email{hangyav@cis.lmu.de}

\author[1]{\fnm{Iryna} \sur{Gurevych}}\email{www.ukp.tu-darmstadt.de}

\author[2]{\fnm{Alexander} \sur{Fraser}}\email{fraser@cis.lmu.de}

\affil[1]{\orgdiv{Ubiquitous Knowledge Processing Lab (UKP Lab)}, \orgname{Department of Computer Science}, \orgaddress{Technical University of Darmstadt}}

\affil[2]{\orgdiv{Center for Information and Language Processing}, \orgname{LMU Munich}}

\abstract{
The goal of hate speech detection is to filter negative online content aiming at certain groups of people.
Due to the easy accessibility of social media platforms it is crucial to protect everyone which requires building hate speech detection systems for a wide range of languages.
However, the available labeled hate speech datasets are limited making it problematic to build systems for many languages.
In this paper we focus on cross-lingual transfer learning to support hate speech detection in low-resource languages. We leverage cross-lingual word embeddings to train our neural network systems on the source language and apply it to the target language, which lacks labeled examples, and show that good performance can be achieved. We then incorporate unlabeled target language data for further model improvements by bootstrapping labels using an ensemble of different model architectures. Furthermore, we investigate the issue of label imbalance of hate speech datasets, since the high ratio of non-hate examples compared to hate examples often leads to low model performance. We test simple data undersampling and oversampling techniques and show their effectiveness.

}


\maketitle

\section{Introduction}\label{sec1}

Due to the increased digitization of society, the impact of online discourse on everyday life is becoming more pronounced. A single hateful message shared on social media now has the potential to incite violent offline movements, as well as exert a negative emotional impact on millions of readers. For this reason, platforms such as Twitter and Facebook have created community policies to ensure civil conduct on the part of their users. The goal is to filter hate speech, which unlike mere offensive or vulgar content, is exclusively designed to attack or denigrate entire groups of people and is generally agreed to add no constructive value to discussions.
But with the sheer amount of posts being published, it is becoming difficult for humans to moderate them in a complete and timely manner. Different annotators are also not guaranteed to agree on every classification, even in the presence of well-defined annotation guidelines. Moreover, due to their repeated and prolonged exposure to negative content, many moderators experience a decline in mental health \cite{vidgen}. For these reasons, automatic hate speech detection has become a field of high interest.

In general, the task of classifying hate speech has been acknowledged as difficult \cite{gibert}. One reason is data scarcity: there are currently few public hate speech datasets available, and the majority of them are for English. Thus, building systems for lower-resource languages is even more challenging \cite{vidgen}.
An additional difficulty of the task is the need to precisely define hate speech. While many people have an intuitive understanding of what hate speech is, this does not easily translate to a finite set of characteristics. Different hate speech datasets often deal with specific hate speech subtypes,
such as hate-speech only against refugees, women or certain nationalities, leading to stark differences between the content of their hate speech classes and making the available resources for a given set of hate speech subtypes in a low-resource language even scarcer. 

The goal of this paper is to reduce these difficulties by exploiting available resources from other languages. We address data scarcity in German, a generally high-resource language but a language for which there are not yet many hate speech datasets available (only 4 labeled corpora compared to the 25 available for English \cite{vidgen}). Our method is applied in a zero-shot setup that assumes no annotated training data in German. We develop a cross-lingual transfer learning approach based on cross-lingual word embeddings (CLWEs) and neural classifiers to provide access to hate speech data in English. We rely on a widely used English dataset
\cite{gibert} as our source language data and the German dataset of the GermEval Shared Task on the Identification of Offensive Language \cite{germevalproc} as our target language data in our experiments. However as we discuss later, their annotation had to be modified using a few simple rules to ensure label compatibility.

In addition to training only on English, we leverage further data to improve our systems.
Towards this end, we bootstrap on two unlabeled German datasets, one of which we crawled from the web. Using an ensemble of our cross-lingual models we predict the labels of previously-unseen data and assign labels with majority voting. We then use this bootstrapped data to further fine-tune the English-trained models. We find that for the majority of our architectures, cross-lingual performance after fine-tuning improves scores within the hate speech class as well as macro-average scores.

Since the majority of social media content is non-hateful, the datasets' label distributions are skewed towards the no-hate label which often leads to training issues, especially in case of small training corpora. For this reason we perform a series of additional experiments to test the impact of class ratio on model performance. We create several over- and undersampled versions of our training sets and compare the models' performance. Our results suggest that severe class imbalance is indeed a problem, but that the best method to overcome it depends on the dataset size.

Our presented approaches are language-independent due to the use of widely available CLWEs making them applicable to many languages.
We show that good performance can be achieved by using only English training data which can be further improved by leveraging unlabeled German sentences as well.
We show that show that class imbalance of hate speech datasets have to be compensated with sampling techniques.

\section{Previous Work}\label{sec2}

\subsection{What is Hate Speech?}\label{subsec1}

Hate speech detection is typically treated as a subtype of offensive language detection. However, \cite{davidson} note that the two concepts are often wrongly conflated due to a lack of a precise definition of hate speech. Other terms that are frequently used in the context of this task are \enquote{abuse} and \enquote{toxicity}, but depending on the dataset these might be treated as synonyms for hate speech or as different categories entirely. The lack of a common consistent practice acts to blur the distinction between hate speech and its possible subtypes, posing a challenge for this research task \cite{fortuna2020}.

Hate speech datasets also differ in annotation schema. This reflects the multifaceted nature of hate speech, as it can be directed against individuals or groups, and can have varying themes such as race, gender, or disability. Thus there are datasets whose annotation schemas distinguish between racism and sexism, as well as datasets specific to certain target groups. The dataset of \cite{bretschneider} views hate speech as \enquote{offensive} statements that express \enquote{fear and aggression}, and collects statements of this nature that are directed against foreigners, while hate speech exclusively against refugees and Muslims are the focus of \cite{bjornross}. The dataset of \cite{davidson} defines hate speech as a statement that \enquote{expresses hatred towards a targeted group or is intended to be derogatory, to humiliate or to insult members of the group}. The three datasets of \cite{hasoc} do not focus on one particular target and contain a diverse set of sentences labeled as hateful.

Tables \ref{hatevariations} illustrates the noticeable differences in how datasets assign the label of hate speech.\footnote{The data samples in this paper are shown for explanatory purposes and do not represent the views of the authors.} Sentences 1 and 2 both direct vulgar language and imagery at politicians, however only Sentence 1 was labeled as hateful in its dataset. Sentences 3 and 4 are both prejudiced against people of a certain nationality and expresses the desire for those people to be removed from somewhere. However, only Sentence 3 was labeled as hate speech. Sentences 5 and 6 both ascribe negative qualities to a race of people, however only Sentence 5 was labeled as hate speech.

\begin{table}[t]
\setcitestyle{numbers}
\begin{center}
\resizebox{0.99\textwidth}{!}{
    \begin{tabular}{llrc}
    \toprule & \textbf{Sentence}  & \textbf{Label} & \textbf{Dataset} \\ 
    \midrule
    1. & Your momma should've swallowed...  Trump, the poster child for & Hate & \cite{hasoc}\\ 
    & retroactive abortions.  \#FuckTrump & \\
    2. & \#Merkel Wie ist diese Schlampe eigentlich auf die Idee &  noHate & \cite{bjornross}\\ 
    & gekommen das Land so tiefgreifend zu verändern? \\
    & \#Merkelmussweg \#refugeesnotwelcome . & \\
    & \footnotesize{\textit{en.\#Merkel How did this slut ever get the idea to change the }}\\
    & \footnotesize{\textit{country so radically? \#Merkelmustgo \#refugeesnotwelcome}} & \\
    \midrule
    3. & Mexicans have plenty of open space in their countries to develop & Hate & \cite{gibert} \\
    & they don't need ours. & \\
    4. & Wir wollen keine Russen hier! & noHate & \cite{bretschneider} \\
    & \footnotesize{\textit{en. We don't want any Russians here!}}\\
    \midrule
    5. & \#Dutch farmers are white trash. & Hate & \cite{davidson} \\
    6. & Also, those stupid white women who watch Oprah & noHate & \cite{gibert} \\
    & and not realize the fact that Oprah is very racist . & \\
    \botrule
    \end{tabular}
}
\caption{Sentences of similar type carrying different class labels in different datasets. German examples are translated to the best of the authors' ability.}\label{hatevariations}

\end{center}
\end{table}

When working with a single dataset, conforming to its hate speech definition is acceptable and possibly even desirable. In a cross-lingual setup however, where both a source- and target-language dataset are required, class label inconsistencies across datasets have a restricting effect. Depending on the hate speech definition of the source-language dataset, many or all available target-language datasets could be incompatible for use alongside it.
Combined with this is the difficulty that non-English languages have very few resources for hate speech detection. A comprehensive online catalogue published by \cite{vidgen} currently catalogues 25 datasets for English, with a grand total of 800,000 data samples.\footnote{\url{https://hatespeechdata.com}} Arabic, the next-best represented language, has just over 48,000 total examples. German has four datasets totalling around 19,500 samples, found in: \cite{bretschneider}, \cite{bjornross}, \cite{germeval}, and \cite{hasoc}.
In our experiments we apply simple rules to make the selected source and target language datasets compatible for the cross-lingual evaluation.

A sub-aspect to this problem is that hate speech is the minority class in most of these datasets \cite{vidgen}. The dataset of \cite{waseem-hovy} has been observed to consist of 68\% non-hate examples \cite{fortuna2020}. This leads to an even smaller amount of available positive examples for the detection of hate speech.

We explore simple under- and oversampling techniques with various label ratios to show the importance of handling the skewed labeled distribution of hate speech datasets.

\subsection{Low-Resource Approaches}\label{subsec2}

CNNs, RNNs, and transformers are the most commonly-used models for hate speech detection and offensive language detection in general \cite{alw1, alw2, alw3, germevalproc, germeval2019}. Two architectures from the 2018 GermEval shared task are particularly relevant to this paper. \cite{fosil} used a CNN following \cite{kim}, while a combination of CNN and BiLSTMs architectures were used to achieve second-best and best performance in the two subtasks respectively \cite{uhhltpaper}. With the increased popularity of transformer-based architectures, models such as BERT \cite{Devlin2019} have successfully been applied to the task as well. A notable example is the 2019 iteration of the GermEval shared task, where the teams using fine-tuned BERT consistently placed among the top performers \cite{germeval2019}. The classifiers we use are based on the three aforementioned architectures.

Data scarcity remains a relevant problem for hate speech detection in non-English languages, and as with many low-resource NLP tasks a common method for achieving good performance is to leverage data from higher-resource languages. This technique is known as cross-lingual transfer learning
which relies on shared representations of languages
in order for knowledge in a source language to be transferable to the target language.
One form of transfer is machine translation, in which the target language data is automatically translated into the source language before classification. However, translation models are prone to producing incorrect translations and require the presence of parallel data to train. A more efficient method to achieve cross-lingual transfer with machine learning models such as CNNs and LSTMs is to use Cross-Lingual Word Embeddings (CLWEs).

Word embeddings provide a means of representing words numerically, thus making important linguistic properties such as semantic similarity accessible to machines. Popular methods are
founded upon the idea that semantically-similar words such as \enquote{joyful} and \enquote{happy} occur in similar contexts \cite{Mikolov2013,bojanowski,Devlin2019}. In a cross-lingual NLP task, word embeddings for both the source and target language are needed
which are aligned, i.e., the vector of a word in the source language is similar to that of its target-language translation. As a result, a source-language sentence is represented with a similar set of vectors as its translations, thus a model trained on the source language may be applied to the target language without any intermediate steps. Various approaches were proposed to build CLWEs, such as the methods based on the idea of mapping independent monolingual embeddings to a shared vector space \cite{Mikolov2013BWE,Conneau,Artetxe2018} or the approaches learning such spaces jointly \cite{Devlin2019}. In our work we rely on both types of approaches, more precisely we use the MUSE \cite{Conneau} and the multilingual BERT \cite{Devlin2019} models.

Cross-lingual transfer techniques were applied for hate speech detection in
\cite{ranasinghe2020} by training transformer-based architectures on English data and using the learned weights to initialize models which are trained on target language data for improved performance. Similarly, a small number of target language samples were concatenated with the source language training data in \cite{stappen2020}.
In \cite{udsw} bilingual word embeddings were used to leverage additional source language data by augmenting the available German training data with English labeled samples.
\cite{hinglish} utilize a cross-lingual transfer procedure for hate speech detection in Hinglish, a code-switched language that uses both Hindi and English words. By first training a CNN and an LSTM on an English dataset, then fine-tuning the models on Hinglish, better performance was achieved compared to a Hinglish-only model. However, this work relied on having labeled data for the target language. In contrast, our approach requires no target language annotations.

Parallel to cross-lingual transfer learning
\cite{bootstrapping} present a bootstrapping-based approach that annotates new data for named entity recognition to improve the performance in low-resource scenarios. First a set of classifiers are trained, which are then applied to an unlabeled set with majority voting. The extended corpus is used to improve the performance by retraining the models from scratch. For hate speech detection, \cite{bigoulaeva} combined the bootstrapping procedure of \cite{bootstrapping} with the fine-tuning procedure of \cite{hinglish} by first bootstrapping German-language hate speech data then using it to fine-tune CNN and BiLSTM classifiers. This resulted in improved performance for both architectures. In this work we follow \cite{bigoulaeva}, additionally using mBERT alongside the CNN and BiLSTM.

\subsection{Imbalanced Classes}\label{subsec3}

Hate speech datasets have an inherent tendency to be imbalanced and this problem must also be addressed in the cross-lingual setup. Even if there is a greater amount of data available for the source language than for the target language, the source-language data may have too few minority class samples to enable good performance. In order to take full advantage of cross-lingual training data, measures must be taken to compensate for any class imbalance of the source-language dataset and learn appropriately from the minority class.

\cite{johnsonsurvey} differentiate between \textit{data-level} and \textit{algorithm-level} methods for dealing with class imbalance. The former is concerned with influencing the data distribution directly through over- or undersampling the data items. The latter is concerned with adjusting model behavior during training by means of cost-sensitive training, selecting certain loss functions, and altering output thresholds. \textit{Hybrid} methods also exist which combine both data-level and algorithm-level techniques.

Due to their simplicity we explore over- and undersampling techniques in our work.
They respectively involve duplicating random samples from the minority class and removing random samples from the majority class. Previous research with feature-based machine learning models suggests that oversampling delivers slightly better performance than undersampling, likely because undersampling removes data \cite{osussurvey, haua}. We test the efficacy of over- and undersampling hate speech datasets on our neural networks.

\section{Experimental Setup}
\label{sec3}

This section introduces the setup of our experiments. First we discuss our chosen datasets, showing their class distributions and the degree of overlap in hate speech definition. Finally we present our three models based on CNN, BiLSTM, and mBERT architectures respectively.

\subsection{Datasets}\label{subsec31}

To ensure the validity of our cross-lingual setup it was necessary to choose a source- and target-language dataset pair such that the hate speech classes of the two overlapped. Despite English being a high-resource language, choosing a dataset with a narrow-focused hate speech definition would potentially limit the number of German datasets that could be used for testing. For this reason we sought out an English dataset with a broad hate speech definition, since it would more likely be compatible with the available German datasets.

\begin{table}[t]
\setcitestyle{numbers}
\centering
\resizebox{0.99\textwidth}{!}{
    \begin{tabular}{lll}
    \toprule & \textbf{Sentence}  & \textbf{Label} \\ \midrule
    1. & This film tells the story of a poor victimised African boy & noHate \\ 
    & (Joseph) who was allowed into Ireland. & \\
    2. & There are London areas that could be described as nothing less & Hate \\ 
    &  than little islamic republics. & \\
    3. & But unfortunately, Maine's become the dumping ground & Hate \\ 
    & for Somalis and other African trash. & \\
    4. & It is nothing short of hysterical that you are such a retard you  & noHate \\
    & have to TELL us in your username that you are  educated. & \\
    5. & im sorry about being a bitch.i was just hurt.hope you find your talk & noHate \\ 
    & with someone. ifyouwant to talk to someone who is trying to get their & \\
    & degrees pm me on sf or im me on aol or yahoo & \\
    \botrule
    \end{tabular}
}
\caption{Sample hate and non-hate comments from the Stormfront dataset \cite{gibert}.}
\label{giberthnh}
\end{table}

One such English dataset is found in \cite{gibert} who define hate speech as \enquote{a deliberate attack directed towards a specific group of people motivated by aspects of the group’s identity}. This dataset features text scraped from the white-nationalist forum Stormfront and will be referred to as the Stormfront dataset. Due to its broad hate speech definition and its decent size (ca. 10,000 examples), it was chosen as the training set for this paper.
Table \ref{giberthnh} illustrates some \enquote*{Hate} and \enquote*{noHate} sentences from the Stormfront dataset. Sentence 1 is not an example of hate speech, since it has a neutral sentiment and does not ascribe the qualities \enquote*{poor} and \enquote*{victimized} to an entire group of people. Sentences 2 and 3 are examples of hate speech directed at religious and racial groups, respectively. Sentence 4 is an attack on an individual that uses the derogatory term \enquote{retard} to ascribe low intelligence, but was assigned the \enquote*{noHate} label since it did not address a group. Finally, Sentence 5 uses the profane and derogatory word \enquote{bitch} in a non-attacking context.

Our choice for the target dataset was the dataset of German-language tweets presented with the 2018 GermEval Shared Task on the Identification of Offensive Language \cite{germevalproc}. The shared task focused on the detection of offensive language in general (the coarse-grained task), along with the detection of three of its subtypes (the fine-grained task): \enquote*{Insult}, \enquote*{Profanity}, and \enquote*{Abuse}. A tweet is assigned the \enquote*{Abuse} label if \enquote{... the target of judgment is seen as a representative of a group and it is ascribed negative qualities that are taken to be universal, omnipresent and unchangeable characteristics of the group} \cite{germevalproc}. Importantly, this definition keeps the nature of the target group general and is therefore compatible with the hate speech definition in \cite{gibert}. Thus, the GermEval dataset was used as our test set for the cross-lingual experiments.
However labels had to be aligned with the Stormfront dataset which we discuss in Section~\ref{annodiscrep}.

\begin{table}[t]
\setcitestyle{numbers}
\centering
\resizebox{0.99\textwidth}{!}{
    \begin{tabular}{lll}
    \toprule & \textbf{Sentence}  & \textbf{Label} \\ \midrule
    1. & @ShakRiet @Heinrich\_Krug So ist es....wir haben Maria vergessen...& Other \\ 
    & als hätte sie nie existiert....schämt euch...! & \\
    & \footnotesize{\textit{en. That's how it is... we have forgotten Maria...as if she never existed...}} & \\
    & \footnotesize{\textit{shame on you...!}} \\
    2. & Martin Schulz ist 2x sitzen geblieben und hat keinen Schulabschluss. & Insult \\
    &  Wie kann denn so ein Nulltipper als Kanzlerkandidat aufgestellt werden? & \\
    & \footnotesize{\textit{en. Martin Schulz was held back in school twice and has no diploma.}}  & \\
    & \footnotesize{\textit{How can that kind of idiot be held for a chancellor candidate?}}  & \\
    3. & Wir sollten den deutschen Kinder und Frauen gedenken die durch den & Abuse \\
    & \#Islam ermordet wurden. & \\
    & \footnotesize{\textit{en. We should commemorate the German children and women murdered}} & \\
    & \footnotesize{\textit{by \#Islam.}}  & \\
    4. & @HenHoffgaard @mboe0407 Da die Kirche jeher den Herrschenden in den & Profanity \\  
    & Arsch gekrochen ist, inkl. Hitler, wundert es mich nicht & \\
    & \footnotesize{\textit{en. Well since the church always kissed the ass of the ruling elite,}} & \\  & \footnotesize{\textit{including Hitler, this doesn't surprise me.}} & \\
    5. & @Nacktmagazin @DuHugonotte Und zum Nachtisch einen Mohrenkopf .   & Abuse \\
     & \footnotesize{\textit{en. And for dessert a Mohrenkopf (head of a Moor / a kind of candy)}} \\
    \botrule
    \end{tabular}
}
\caption{Sample comments from the GermEval dataset \cite{germeval}.}
\label{germevalhnh}
\end{table}

Table \ref{germevalhnh} shows samples of various classes from the GermEval dataset. Sentence 1 expresses negative emotions about a specific person being forgotten but does not seek to attack or denigrate anyone. Sentence 2 insults a single politician with a nickname \enquote{Nulltipper} \textit{(en. \enquote{idiot})} and the lack of a school diploma to ascribe low intelligence. Sentence 3 is an example of \enquote{Abuse}, since it ascribes acts of murder to an entire religious group. Sentence 4 is an example of the \enquote{Profanity} category as it contains the profane phrase \enquote{in den Arsch gekrochen}, while not being verbosely critical or attacking. Finally, Sentence 5 is another example of the \enquote{Abuse} class, since it uses the term \enquote{Mohrenkopf}, which typically denotes a kind of candy, as a derogatory designation for dark-skinned individuals.

\subsubsection{Annotation Discrepancies}\label{annodiscrep}

Examining the two datasets' hate speech definitions and labeled hate speech examples in Tables \ref{giberthnh} and \ref{germevalhnh}, it is clear that GermEval's \enquote{Abuse} category corresponds with the \enquote*{Hate} label of the Stormfront dataset. However, the differing annotation taxonomies as well as the different names attached to the compatible categories pose problems for machine learning models, which will expect consistent annotations between training and testing. Therefore it was necessary to make a few simple adjustments to the datasets before beginning our experiments.

The Stormfront dataset's distinction between \enquote*{Hate} and \enquote*{noHate} is an example of a binary annotation schema. Additionally the dataset contains a \enquote*{Relation} label for sentences that had to be considered in context with others to acquire a hateful meaning, and a \enquote*{Skip} label for when the sentence was either non-English or not meaningful enough to be given either of the binary labels. In contrast, the GermEval dataset features a two-tiered annotation schema: each tweet carries a label for the coarse-grained task of \enquote*{Offense} vs \enquote*{Other} as well as a fine-grained label that specifies the subtype of offensiveness: either \enquote*{Insult}, \enquote*{Profanity}, or \enquote*{Abuse}. 

To ensure compatibility between these two datasets, we made certain modifications to their labeling schemas that were motivated by the datasets' specific class definitions.
First we simplified the annotation schema of the fine-grained GermEval data into a binary schema. As per the above discussion in Section \ref{subsec31}, we took GermEval's \enquote*{Abuse} label to be the counterpart of the Stormfront dataset's \enquote*{Hate}, relabeling comments belonging to the \enquote*{Other}, \enquote*{Insult}, and \enquote*{Profanity} classes as \enquote*{noHate}. 
Next, we relabeled all \enquote*{Skip} and \enquote*{Relation} samples from the Stormfront dataset to conform with the binary schema. The 92 comments that carried the label \enquote*{Skip}, indicating that they were either non-English or not informative, were relabeled as \enquote*{noHate}. The 168 instances of the \enquote*{Relation} class were relabeled as \enquote*{Hate}, since these sentences were always hateful when placed in context.

After relabeling was completed, we split both datasets into training, development, and test sets. From the Stormfront dataset we form our EN-TEST set by selecting random \enquote*{Hate} and \enquote*{noHate} samples, with a class ratio that roughly reflects the data distribution. We kept the size of this dataset small in the interest of preserving resources for training. Next we draw an equal amount of \enquote*{Hate} and \enquote*{noHate} samples that did not overlap with EN-TEST for our EN-DEV dataset. The remaining samples formed EN-TRAIN.

For the split-up of our GermEval dataset, we follow the work of \cite{uhhltpaper}. The GermEval shared task data comes with an official train and test dataset, the latter of which we keep and name DE-TEST. For our train/dev split, we transfer the last 809 samples from the provided training set to a new development set named DE-DEV for hyperparameter tuning. The remaining samples formed our DE-TRAIN dataset, which will be used only in the bootstrapping experiments. 
Table \ref{traindevtest} shows the class distribution of the resulting datasets. These will form the basis of our experiments. See Tables \ref{sfinit} and \ref{deinit} to compare to the original, unmodified versions of the datasets.

\begin{table}[t]
\centering
\begin{tabular}{@{} lrrc @{}}
\hline
& \textbf{noHate} & \textbf{Hate} & \textbf{Ratio (approx.)} \\\hline
EN-TRAIN &  9,018 & 1,281 & 7:1 \\
EN-DEV & 134 & 20 & 7:1 \\
EN-TEST & 427 & 63 & 7:1 \\ \hline
DE-TRAIN & 3,345 & 855 & 4:1 \\
DE-DEV  & 642 & 167 & 4:1 \\ 
DE-TEST  & 2,759 & 773 & 4:1 \\
\hline
\end{tabular}
\caption{Class distributions of the English and German datasets after relabeling and train/dev splitting.}
\label{traindevtest}
\end{table}

\begin{table}[t]
\centering
\begin{tabular}{@{} lrrrr @{}}
\hline
& \textbf{noHate} & \textbf{Hate} & \textbf{Relation} & \textbf{Skip} \\ \hline
Stormfront & 9,488   & 1,196 & 168 & 92 \\ 
\hline
\end{tabular}
\caption{Original Stormfront dataset before relabeling and train/dev splitting.}
\label{sfinit}
\end{table}

\begin{table}[t]
\centering
\begin{tabular}{@{} lrrrr @{}}
\hline
& \textbf{Other} & \textbf{Abuse} & \textbf{Insult} & \textbf{Prof.} \\ \hline
Train & 3,321 & 1,022 & 595 & 71 \\ 
Test  & 2,330 & 773 & 381 & 48 \\  
\hline
\end{tabular}
\caption{Original GermEval datasets before relabeling and dev splitting from the training set. These were the datasets provided to the shared task participants.}
\label{deinit}
\end{table}

\subsubsection{Addressing Class Imbalance}\label{addressingclassimb}

After the relabeling and train/dev splitting process was complete, we addressed the imbalanced class distributions of the training datasets. Examining Table \ref{traindevtest}, it is clear that there is a greater abundance of \enquote*{noHate} compared to \enquote*{Hate}. This reflects the real-life pattern of hate speech occurring less commonly than regular text. But this poses difficulties for machine learning models, which need plenty of data from both classes in order to be able to generalize \cite{madukwe-etal-2020-data, vidgen}.

Previous research suggests that oversampling the underrepresented class yields good model performance \cite{haua}. In the interest of using all available \enquote*{noHate} data, we manually duplicate the \enquote*{Hate} examples and some \enquote*{noHate} samples from EN-TRAIN to produce a balanced, oversampled dataset named EN-OS[1:1]. The balanced 1:1 class ratio represents the best-case scenario where neither class is in the minority. The resulting dataset is shown in Table \ref{enos}.
We experiment with additional label distributions in Section~\ref{sec5}.

\begin{table}[t]
\centering
\begin{tabular}{@{} lrrc @{}}
\hline
& \textbf{noHate} & \textbf{Hate} & \textbf{Ratio (approx.)} \\\hline
EN-TRAIN &  9,018 & 1,281 & 7:1 \\
EN-OS[1:1] & 9,018  & 9,018 & 1:1 \\
\hline
\end{tabular}
\caption{The unmodified EN-TRAIN dataset and its balanced oversampled version: EN-OS[1:1].}
\label{enos}
\end{table}

\subsection{Models}\label{subsec32}

In our experiments we focus on evaluating neural network architectures, using
monolingual models that have been popularly applied to the task in the past. Our first model is a CNN classifier following \cite{kim}. This model accepts an embedding layer as an input and feeds it into a convolution layer with a variable number of filters. Global max-pooling is performed on the convolution output, and the result is passed through a dense layer. The input word embeddings can either be randomly-initialized, pre-loaded from an outside source, or fine-tuned during training. Optionally, two of these variants may be used at the same time. We used only one form of embeddings, namely our CLWEs, and did not update them during training.
For the remaining model hyperparameters, we used the default values.\footnote{\url{https://github.com/yoonkim/CNN_sentence/blob/master}}

To produce our CLWEs, monolingual embeddings were first trained using FastText SkipGram \cite{fasttext} over English and German NewsCrawl corpora \cite{newscrawl} which contain text dating from 2007 to 2013 and were preprocessed with Moses tools \cite{koehn2006open}. The resulting embeddings were mapped with MUSE \cite{muse}. We used the default parameters of the above mentioned tools.

Our second model is based on the neural model of one of the participants of the 2018 GermEval Shared Task \cite{uhhltpaper}, with some modifications for compatibility with our cross-lingual setup. In our version, an input layer of our CLWEs was fed into a BiLSTM layer of 100 units. The output of this BiLSTM layer was then fed into a convolution layer with three feature maps of 200 units each, with respective kernel sizes of 3, 4, and 5. Global max-pooling was applied after each convolution, and the output of this step was fed to a dense layer of 100 units.

Our third architecture is multilingual BERT, which was pre-trained on Wikipedia data from 104 languages \cite{Devlin2019}. This architecture has the advantage of not needing CLWEs as a resource and can be tuned and tested on a source and target language directly. For the sake of consistency in discussions about the other two architectures, we will henceforth refer to the process of tuning mBERT as \enquote{training}.

\section{Results}
\label{sec6}

We conduct our cross-lingual experiments by training the three architectures from Section \ref{subsec32} on English and testing on German. We use our EN-OS[1:1] dataset for training. Since the testing language was German, hyperparameters such as epoch count, learning rate, and class weights were optimized on DE-DEV.

Table \ref{crosslingualperf} shows the performance of these models when tested on DE-TEST. All three models manage to transfer their knowledge of \enquote*{noHate} from English to German, with the CNN and mBERT in particular achieving classwise \enquote*{Hate} scores greater than 50 points: 67.44 precision for the CNN and 52.29 recall for mBERT, respectively. Scores were significantly higher in the \enquote*{noHate} class: The CNN achieved 78.82 precision and mBERT  achieved scores above 60. The BiLSTM had the highest performance in \enquote*{noHate}, with precision, recall, and F1 scores all above 75. This is notable since we did not use German-language data at any point. The macro-average scores of the CNN and BiLSTM were relatively tied, however the BiLSTM achieved a macro-average F1 score that was nearly as high as that of mBERT. mBERT's macro-average scores were the highest among the three models. These results show that cross-lingual training with neural networks is a viable option even when no target-language data is available.
These three models will form the ensemble used in Section \ref{sec7}.

Table \ref{enoshyper} shows the hyperparameters that gave optimal performance on EN-OS[1:1]. We observed that mBERT preferred small batch sizes, its scores slightly dropping as batch size was increased. The CNN and BiLSTM in contrast preferred much larger batch sizes and learning rates and exhibited poorer performance when the batch size was lowered. Class weight ratios implemented into the loss function were a relevant parameter for the CNN, which required a slightly greater weight for the \enquote*{noHate} class. Despite this measure the CNN exhibited severe overfitting behavior, becoming skewed towards predicting only one of the two class labels as epoch count increased. Notably this pattern persisted despite class weight and learning rate tuning. Training on a single epoch with a large batch size yielded optimal performance.

\begin{table}[t]
\centering
\resizebox{0.99\textwidth}{!}{
\begin{tabular}{ccccccccccc}
\hline
\multicolumn{1}{c}{\textbf{Model}} & \multicolumn{1}{c}{\textbf{Accuracy}} & \multicolumn{3}{c}{\textbf{noHate}} & \multicolumn{3}{c}{\textbf{Hate}} & \multicolumn{3}{c}{\textbf{Macro-Avg}} \\ 
\hline
&  & \textbf{P} & \textbf{R} & \textbf{F1} & \textbf{P} & \textbf{R} & \textbf{F1} & \textbf{P} & \textbf{R} & \textbf{F1} \\
CNN & 40.91	&78.82	&33.56	&47.07	&21.94	&67.44	&33.11	&50.38	&50.50	&40.09 \\
BiLSTM & 70.44	&77.80	&86.99	&82.14	&19.69	&11.38	&14.43	&48.74	&49.19	&48.28
\\
mBERT & 66.39	&67.83	&93.30	&78.55	&52.29	&14.23	&22.37	&60.06	&53.77	& 50.46 \\
\hline
\end{tabular}
}
\caption{Model performance on DE-TEST after training on EN-OS[1:1].}
\label{crosslingualperf}
\end{table}

\begin{table}[t] 
\centering
\begin{tabular}{@{} lrrrrrrr @{}}
\hline
& \textbf{noHate} & \textbf{Hate} & \textbf{Dropout} & \textbf{Learn Rate} & \textbf{Batch Size} & \textbf{Epochs}\\ \hline
CNN &  0.6	& 0.4 & 0.7 &  $10^{-4}$ & 50 & 1 \\ 
BiLSTM   & 0.5& 0.5& 0.2 & $30^{-3}$ & 40 & 30   \\ 
mBERT  &  -- & -- & 0.2 & $10^{-5}$ & 5 & 10\\ 
\hline
\end{tabular}
\caption{Optimal hyperparameters for training on EN-OS[1:1]. The first two columns represent class weights, which were not implemented for mBERT.}
\label{enoshyper}
\end{table}

\subsection{Bootstrapping}
\label{sec7}
Although cross-lingual transfer learning techniques are applicable to zero-shot hate speech detection, the discussed data scarcity issues, such as low amount of positive hate speech labeled examples, hinder the performance. To mitigate these issues, this phase of cross-lingual experiments is centered around data augmentation and fine-tuning. For this we relied on two target language unlabeled datasets which we labeled automatically using an ensemble-based approach following \cite{bigoulaeva}. Our relabeling ensemble consisted of the three neural models in Table \ref{crosslingualperf}. We test these models on two sources of German data: the DE-TRAIN dataset (See Table \ref{traindevtest}) and the DE-NEW dataset to be detailed in Section \ref{bsonsf}. For each of the two datasets, we applied
all three of our models and assigned a final label to each sentence based on majority voting.

For each bootstrapping dataset we take the three models from Table \ref{crosslingualperf}, which had originally been trained on EN-OS[1:1], and resumed their training on the bootstrapping dataset, using altered hyperparameter settings as needed to optimize performance. We then test the performance of the fine-tuned models on DE-TEST.

\subsubsection{Bootstrapping on DE-TRAIN}\label{bsondetrain}

In this first phase of the bootstrapping experiments, we apply our ensemble to the DE-TRAIN dataset and collect the majority-vote classification results into a new dataset called DE-REL*.
We simulate DE-TRAIN as an unlabeled dataset, since it was not used for training of our models.

Table \ref{derelstar} shows the confusion matrix for the labels of DE-REL*. It is clear that this dataset consists predominantly of \enquote*{noHate} examples, with a severely imbalanced ratio of 43:1. 573 true \enquote*{Hate} examples were mistakenly labeled by the ensemble as \enquote*{noHate}, while 42 true \enquote*{noHate} examples were mistakenly labeled as \enquote*{Hate}. Proportionally more classification errors were made in the \enquote*{Hate} class, reflecting the models' higher precision, recall, and F1 scores in \enquote*{noHate} as can be seen from Table \ref{crosslingualperf}. 

\paragraph{The Labels of DE-REL*}
\label{labelsofderel}

\begin{table}
\centering
\begin{tabular}{@{} l|rr @{}}
\hline
& noHate & Hate \\ \hline
noHate & 2,688 & 42 \\
Hate & 573 & 34 \\
\hline
Total  & 3,261  & 76  \\
\hline
\end{tabular}
\caption{
Confusion matrix of the ensemble-relabeled DE-REL* compared to the original annotations in DE-TRAIN. Gold and predicted labels are shown in the rows and columns respectively.
}
\label{derelstar}
\end{table}

Table \ref{ensemblelabels} provides a closer look at some correctly- and incorrectly-classified examples from DE-REL*, as compared to the original gold labels of DE-TRAIN. Sentence 1 was correctly labeled by the ensemble as \enquote*{Hate}, as it attributes negative qualities such as violence to a religious group. Sentence 2 was also correctly classified as hate speech, as it expresses approval of prejudiced actions towards people with brown skin. Sentence 3 was correctly recognized as \enquote*{noHate}, although it contains a potentially contentious word \enquote*{Hetze} (\textit{en. \enquote*{hate, agitation}}), which often occurs in contexts of hate speech. This indicates that the ensemble has some knowledge of hate speech features that go beyond lexical cues. Finally, Sentence 4 was falsely labeled by the ensemble as \enquote*{noHate}. This was likely a challenging example for the ensemble due to it being a form of gender-related hate speech that is not abundantly encountered on a white supremacy forum.

\paragraph{Performance}
\label{bs1results}
Table \ref{ft1perf} shows the English-trained models' performance on DE-TEST after fine-tuning on DE-REL*. Both mBERT and the BiLSTM improve their performance in several areas. The BiLSTM's classwise recall and F1 for \enquote*{Hate} increased by 4.14 points and 3.04 points, respectively. Its macro-average F1 increased by 0.49. mBERT's classwise \enquote*{Hate} improvements were more modest, its precision increasing by 1.56 points and its F1 by 1.27. Additionally its macro-average F1 increased by 0.71. The BiLSTM's greater improvements could be due to the model having had too little training data before, while mBERT had already become mostly saturated by the English training data.

The only model to perform worse after fine-tuning was the CNN, which during training outputted either only \enquote*{Hate} or only \enquote*{noHate} predictions. The latter is associated with higher macro-average performance since \enquote*{noHate} is the majority class of DE-TEST. This result is likely due to poor initial training of the CNN. Recalling from Section \ref{sec6}, the CNN was trained on EN-OS[1:1] for only one epoch, as it exhibited overfitting behavior otherwise. It is likely however that this training was suboptimal and that the single epoch of training was not enough for the CNN to sufficiently learn from its training data.

\begin{table}[t]
\centering
\resizebox{0.9\textwidth}{!}{
\begin{tabular}{ccccccccccc}
\hline
\multicolumn{1}{c}{\textbf{Model}} & \multicolumn{1}{c}{\textbf{Accuracy}} & \multicolumn{3}{c}{\textbf{noHate}} & \multicolumn{3}{c}{\textbf{Hate}} & \multicolumn{3}{c}{\textbf{Macro-Avg}} \\ 
\hline
&  & \textbf{P} & \textbf{R} & \textbf{F1} & \textbf{P} & \textbf{R} & \textbf{F1} & \textbf{P} & \textbf{R} & \textbf{F1} \\
CNN & 78.11	&78.11	&100.00	&87.71	&0	&0.00	&0.00	&39.06	&50.00	&43.86 \\
BiLSTM & 67.89	&77.72	&82.57	&80.07	&19.97	&15.52	&17.47	&48.84	&49.05	&48.77 \\
mBERT & 66.70	&68.07	&93.30	&78.71	&53.85	&15.14	&23.64	&60.96	&54.22	&51.17 \\
\hline
\end{tabular}
}
\caption{Model performance on DE-TEST after training on EN-OS and fine-tuning on DE-REL*.}
\label{ft1perf}
\end{table}

Table \ref{derelhyper} shows the hyperparameter settings that were used for fine-tuning on DE-REL*. We observed that tuning the class weights for the CNN as well as the dropout had no effect on the overfitting performance. The BiLSTM however achieved balanced performance with similar hyperparameter settings to those of the CNN. mBERT preferred a smaller batch size and improved performance when fine-tuned for more epochs than the other architectures.

\begin{table}[t] 
\centering
\begin{tabular}{@{} lcccccc @{}}
\hline
& \textbf{noHate} & \textbf{Hate} & \textbf{Dropout} & \textbf{Learn Rate} & \textbf{Batch Size} & \textbf{Epochs}\\ \hline
CNN &  0.01	& 0.99 & 0.2 &  $10^{-6}$ & 30 & 1  \\ 
BiLSTM   & 0.1 & 0.9 & 0.7 & $10^{-6}$ & 50 & 2   \\ 
mBERT  &  -- & -- & 0.5  & $10^{-5}$ & 10 & 10\\ 
\hline
\end{tabular}
\caption{Optimal hyperparameters for fine-tuning on DE-REL*. The first two columns represent class weights, which were not implemented for mBERT.}
\label{derelhyper}
\end{table}

\subsubsection{Bootstrapping on German Stormfront Data}
\label{bsonsf}

In the second bootstrapping experiment, we use the DE-NEW dataset collected by \cite{bigoulaeva}, which was crawled from a German-language thread within the Stormfront forum. Since this data comes from the same website as EN-OS[1:1], there is good stylistic compatibility between the English and German data.

At the time of crawling, the source thread had around 5,500 posts\footnote{The website was crawled using the Python Scrapy library.}. These consisted predominantly of comments written in German, although many were written in English. To account for the typical prevalence of lengthy posts in a forum setting, \cite{bigoulaeva} considered each paragraph distinguished by a newline to be a separate text sample.
Before the data could be used for training, some manual preprocessing was performed to ensure compatibility with the format of a tweet. Table \ref{denewmodifications} shows what texts were kept and removed.
Additionally, the following errors in the texts were manually corrected and kept:

\begin{itemize}
\item \enquote*{tut mir} and \enquote*{leid} $\rightarrow$ \enquote*{tut mir leid}
\item \enquote*{d aß} $\rightarrow$ \enquote*{daß}
\end{itemize}

\begin{table}
\centering
\begin{tabular}{@{} rl|rl @{}}
\hline
 & \textbf{Removed} & & \textbf{Kept} \\ \hline
1. & Non-German text & 1. & Quotes or news article snippets  \\
 &Bullet-point lists &  & under 1000 characters.   \\
2. & Quotes from books, articles, etc. & 2. & Mixed English/German sentences \\
 &over 1000 characters & 3. & Multi-line interview dialogue,  \\
3. & Extremely short lines: names, & & with each line considered  \\
 &one-word responses, timestamps, & & as a distinct text sample. \\
 &letter salutations & 4. & Mixed English/German sentences \\ 
4. & Lines or sentences that were cut off & &  and Anglicisms \\
 &without any clear continuation & \\
\hline
\end{tabular}
\caption{Preprocessing steps for the DE-NEW dataset.}
\label{denewmodifications}
\end{table}

\begin{table}[]
\centering
\resizebox{0.99\textwidth}{!}{
\begin{tabular}{p{.2cm} p{.1cm} p{9.0cm} p{1.5cm} p{1cm}}
\hline
& & \textbf{Sentence} & \textbf{Ensemble} & \textbf{Gold} \\ \hline
\multirow{4}{*}{\rotatebox{90}{DE-REL*~}} & 1. & \#Islamisierung "Zusammenstöße zw GLÄUBIGEN (Richtig: \#Islamisten) \& Sicherheitskräften" - \#Tagesschau sendet bereits im \#Scharia-Modus   & Hate &  Hate \\  
 & & \textit{en. \enquote{Confrontations between BELIEVERS (Actually: \#Islamists) \& the police force} - \#Tagesschau is already broadcasting in \#Sharia-Mode} \\
& 2. & Dortmund haut wieder auf die Kacke ! Die wollen die ganzen Braunhäutigen verjagen mit dem Slogan BUNT STATT BRAUN !! BRAVO !!!  & Hate  & Hate \\
 & & \textit{en. Dortmund is rising up again! They want to drive out all the brown-skinned with the slogan \enquote{COLORFUL INSTEAD OF BROWN!} BRAVO!}  \\
& 3. & @Dora\_Bromberger @lawyerberlin Hetze haben weder Fakten noch Argumente je ersetzt. & noHate & noHate \\ 
 & & \textit{en. Hate has never replaced facts or arguments.}  \\ 
& 4. & @Sammy\_aus\_Cux Von mir aus gibt es dieses intersexuell aber die eigentlichen Geschlechter sind männlein und weiblein & noHate &  Hate \\ 
& & \textit{en. For all I care these intersexuals do exist but the real sexes are male and female} \\
\hline
\multirow{4}{*}{\rotatebox{90}{DE-NEW~}} & 5. & Das Problem mit diesen Mischlingsehen ist, dass diese Maenner sich nicht um Ihre Familie kuemmern und die meisten dann von Sozialhilfe leben. Die Kinder sehen aus wie Orang-Utans, sind nicht sehr intelligent und werden ueberall gehaenselt, was wiederum zu einem kriminellen Lenbenstil fuehrt.	& Hate & (Hate) \\
& & \textit{en. The problem with these mixed marriages is that these men don't care for their families and most of them then live off of welfare. The kids look like orangutans, aren't very intelligent, and get teased everywhere, which then leads to a criminal lifestyle.}  \\
& 6. & Dreckspack verfluchtes! Wie bescheuert kann man sein? & Hate  & (Hate) \\
 & & \textit{en. Cursed pack of scum! How crackbrained can you get?}  \\
& 7. & Ich glaube die Türken warten auf die Ergebnisse der türkischen Experten & noHate & (noHate) \\ 
 & & \textit{en. I think the Turks are waiting for the results of the Turkish experts.}  \\
& 8. & Der in Tutzing in Oberbayern lebende kanadische \textbf{Holocaust-Leugner} Alfred Schaefer ist wegen Volksverhetzung angeklagt. Der 63-Jährige selbst hat den Verhandlungstermin am 4. Mai vor dem Amtsgericht Dresden mit den Worten er sei \enquote{vor die Inquisition geladen} publik gemacht und angekündigt, den Prozess dazu zu nutzen, in langatmiger Form den \textbf{nationalsozialistischen Völkermord} an den Juden in Frage zu stellen. & noHate & (noHate) \\
& & \textit{en. The Canadian \textbf{Holocaust-denier} Alfred Schaefer, who lives in Tutzing in Upper Bavaria, has been charged with sedition. On his part, the 63-year-old made his trial appointment before the court of Dresden on the 4th of May public by saying he had been \enquote{invited by the Inquisition}, and announced that he would use the process to verbosely call into question the \textbf{national-socialist genocide} of the Jewish people.} \\
& 9. & Könnte es sein, jene wenigen Privilegierten beginnen zu begreifen, daß Macht und Reichtum nichts gegen das Streben nach historischer Exaktheit ausrichten können?   & Hate &  (noHate) \\  
 & & \textit{en. Could it be that those few privileged are beginning to realize that power and riches have nothing against the striving towards historical precision?} \\
\hline
\end{tabular}
}
\caption{Correct and incorrect ensemble labels for either DE-REL* or DE-NEW. Gold labels for DE-NEW are given by the authors in brackets.}
\label{ensemblelabels}
\end{table}

As a result of this preprocessing, DE-NEW contains 6,586 text samples, all or nearly all written in German. This dataset was used as the training set during fine-tuning.

Table \ref{denewderel} shows the class distribution of the DE-NEW dataset compared to DE-REL*. DE-NEW is the larger, but interestingly the ensemble's relabeling resulted in both datasets having similar class ratios. This could indicate that the stylistic differences between the Twitter-based text of DE-REL* and the forum-based text of DE-NEW were not a hindering factor for the ensemble.

\begin{table}[t]
\centering
\begin{tabular}{@{} lrrc @{}}
\hline
& \textbf{noHate} & \textbf{Hate} & \textbf{Ratio (approx.)} \\\hline
DE-NEW &  6,437 & 142 & 45:1 \\
DE-REL* & 3,261 & 76 & 43:1 \\
\hline
\end{tabular}
\caption{Class distributions of the two bootstrapped datasets.}
\label{denewderel}
\end{table}

\paragraph{The Labels of DE-NEW}

Since we had no gold labels of DE-NEW to evaluate our ensemble's classifications, we manually examined several examples and judged them strictly according to the points of the hate speech definition in \cite{gibert}. Table \ref{ensemblelabels} shows five classifications made by the ensemble. 

Sentence 5 was correctly identified as \enquote*{Hate}, as it is derogatory towards families of mixed races, employing dehumanizing comparisons and attributing low intelligence. Sentence 6 was also correctly identified as \enquote*{Hate}. Although the target group is unclear, the group is also described with dehumanizing language and is portrayed as being dirty and unintelligent. Sentence 7 is a neutral descriptive statement that does not attack the group of Turkish people, and was correctly recognized as \enquote*{noHate}. Similarly, Example 8 is a neutral descriptive account, despite discussing a figure of controversy and using terminology (shown in bold) that would likely be associated with hateful discourse: \enquote{nationalsozialistischen Völkermord} (\textit{en. National-Socialist/Nazi genocide}), and \enquote{Holocaust-Leugner} (\textit{en. Holocaust-denier}). Together with Sentence 7, this again shows that the ensemble learned more complex features of hate speech than lexical cues (See Section \ref{bsondetrain}).

Sentence 9 was another challenge for the ensemble. It was labeled as \enquote*{Hate} despite not having any telling signs of hate speech, likely due to discourse about privilege, power and riches having occurred elsewhere in the Stormfront data in more hateful contexts. This would lead the models of the ensemble to recognize that these groups are typically ones to be attacked. Nevertheless we judged this sentence to be an example of \enquote*{noHate}, since when the sentence is considered in isolation it does not attack or dehumanize the groups in question.

\paragraph{Performance}

Table \ref{ft2perf} shows the models' performance on DE-TEST after training on EN-OS and fine-tuning on DE-NEW. As in the first fine-tuning experiment, the BiLSTM and mBERT improved their scores over the original versions trained on EN-OS[1:1].
This time the BiLSTM's classwise \enquote*{Hate} scores improved to a lesser degree, with its precision increasing by 0.55 points and its classwise recall and F1 score dropping slightly. Nevertheless this precision value was higher than after fine-tuning on DE-REL*.  All three of its macro-average measures improved as well and were also higher than in the first fine-tuning round (See Table \ref{ft1perf}).
mBERT experienced a slight decrease in macro-average and classwise \enquote*{Hate} scores. The only \enquote*{Hate} score to improve was precision, which increased by 0.30 points. Classwise recall and F1 in \enquote*{noHate} increased while the precision decreased. 

This lesser degree of improvement in \enquote*{Hate} compared to the first fine-tuning experiment could have been caused by DE-NEW's slightly larger ratio of \enquote*{noHate} to \enquote*{Hate} as compared to DE-REL* in Table \ref{denewderel}. As in the previous bootstrapping experiments, the CNN model worsened after fine-tuning, likely due to poor initial training.

Table \ref{denewhyper} shows the hyperparameter settings that were used for fine-tuning on DE-NEW. As before, tuning these hyperparameters did not mitigate the CNN's overfitting performance. The BiLSTM improved with a smaller batch size than in the previous fine-tuning experiment as well as with a lower learning rate and higher dropout. mBERT's improvements in this fine-tuning experiment were also correlated with different hyperparameters, in this case a small batch size, a lower learning rate, and a reduced epoch count. The reason for this behavior could be the differing class ratios between DE-REL* and DE-NEW.

\begin{table}[t]
\centering
\resizebox{0.99\textwidth}{!}{
\begin{tabular}{ccccccccccc}
\hline
\multicolumn{1}{c}{\textbf{Model}} & \multicolumn{1}{c}{\textbf{Accuracy}} & \multicolumn{3}{c}{\textbf{noHate}} & \multicolumn{3}{c}{\textbf{Hate}} & \multicolumn{3}{c}{\textbf{Macro-Avg}} \\ 
\hline
&  & \textbf{P} & \textbf{R} & \textbf{F1} & \textbf{P} & \textbf{R} & \textbf{F1} & \textbf{P} & \textbf{R} & \textbf{F1} \\
CNN & 78.11	&78.11	&100.00	&87.71	&0.00	&0.00	&0.00	&39.06	&50.00	&43.86 \\
BiLSTM &  71.04	&77.89	&87.86	&82.58	&20.24	&11.00	&14.25	&49.07	&49.43	&48.41
 \\
mBERT & 66.31	&67.27	&95.28	&78.86	&52.59	&10.15	&17.02	&59.93	&52.71	&47.94
 \\
\hline
\end{tabular}
}
\caption{Model performance on DE-TEST after training on EN-OS[1:1] and fine-tuning on DE-NEW.}
\label{ft2perf}
\end{table}

\begin{table}[t] 
\centering
\begin{tabular}{@{} lcccccc @{}}
\hline
& \textbf{noHate} & \textbf{Hate} & \textbf{Dropout} & \textbf{Learn Rate} & \textbf{Batch Size} & \textbf{Epochs}\\ \hline
CNN &  0.01	& 0.99  & 0.9  &  $10^{-4}$ & 2 & 1 \\ 
BiLSTM   & 0.1 & 0.9 & 0.9 & $10^{-7}$ & 20  & 1    \\ 
mBERT  &  -- & -- & 0.6  & $10^{-7}$ & 1 & 5\\ 
\hline
\end{tabular}
\caption{Optimal hyperparameters for fine-tuning on DE-NEW. The first two columns represent class weights, which were not implemented for mBERT.}
\label{denewhyper}
\end{table}

\subsection{Class Imbalance}
\label{sec5}

In this section we take a closer look at the effect of class imbalance on model performance in hate speech detection. To keep the focus on the individual datasets, we perform our experiments monolingually, testing on the same language as for training. We tune hyperparameters on the corresponding development sets.

In our cross-lingual experiments in Section \ref{subsec31} we oversampled our English dataset, noting that oversampling had achieved good results in previous research. Nevertheless, duplication of examples also introduces the risk of overfitting as it saturates the class with stylistically-similar examples \cite{johnsonsurvey, osussurvey}. Here our goal is to investigate whether oversampling is optimal, and if so, at which class ratio. 

We observe from Table \ref{traindevtest} that DE-TRAIN not only has a different class ratio than EN-TRAIN but is also much smaller. Therefore to perform as little duplication as possible we select a set of class ratios for sampling that are based around the ratios of these unmodified datasets. The ratios we sample are 7:1 (as in EN-TRAIN), 4:1 (as in DE-TRAIN), and 1:1 (the balanced scenario). The sampled datasets are named with their language code initials appended with either \enquote*{US} if produced by undersampling or \enquote*{OS} if produced by oversampling. We note that we already have an oversampled English dataset, namely EN-OS[1:1] (See Table \ref{enos}). 
To match the 7:1 ratio of \enquote*{noHate} to \enquote*{Hate} in EN-TRAIN we produce an oversampled version of DE-TRAIN called DE-OS[7:1] with a 7:1 class ratio. Next we produce EN-US[2:1] and DE-US[2:1] by removing appropriate amounts of \enquote*{noHate} examples from EN-TRAIN and DE-TRAIN, respectively.
We keep EN-OS[1:1] from our previous experiments and additionally create EN-US[1:1] and DE-US[1:1], which were produced by removing \enquote*{noHate} examples until their number matched the number of \enquote*{Hate} examples in their respective datasets. Finally, DE-OS[1:1] was produced by duplicating the \enquote*{Hate} examples until they match the number of \enquote*{noHate} examples.
Table \ref{sampledensets}

shows label statistics of  the resulting datasets for English and German.

\begin{table}[t]
\centering
\begin{tabular}{@{} lrrc @{}}
\hline
& \textbf{noHate} & \textbf{Hate} & \textbf{Ratio (approx.)} \\\hline
EN-TRAIN &  9,018 & 1,281 & 7:1 \\
EN-US[2:1] & 2,562   & 1,281 & 2:1 \\ 
EN-US[1:1] & 1,281 & 1,281 & 1:1 \\
EN-OS[1:1] & 9,018 & 9,018 & 1:1 \\
\hline
DE-OS[7:1] & 5,985 & 855 & 7:1 \\
DE-US[2:1] & 1,710 & 855 & 2:1 \\
DE-US[1:1] & 855 & 855 & 1:1 \\
DE-OS[1:1] & 3,345 & 3,345 & 1:1 \\
\hline
\end{tabular}
\caption{English and German training datasets used in our monolingual experiments. Sampled datasets were produced from EN-TRAIN and DE-TRAIN respectively.}
\label{sampledensets}
\end{table}

The results of our experiments for the CNN, the BiLSTM and mBERT architectures are presented respectively in Tables \ref{monocnn}, \ref{monolstm} and \ref{monobert}.
The CNN achieved its highest classwise \enquote*{Hate} scores with the EN-OS[1:1] and EN-US[1:1] datasets. Among the German datasets, the CNN achieved its best \enquote*{Hate} F1 on the two balanced datasets and on DE-US[2:1]. Classwise \enquote*{Hate} performance on DE-OS[7:1] was significantly lower. In particular, the CNN achieved noticeably lower \enquote*{Hate} recall on this dataset than on DE-US[2:1] and DE-US[1:1], despite the \enquote*{Hate} precision scores being similar. Since the total amount of \enquote*{Hate} samples in these three datasets was the same (see Table \ref{sampledensets}), the class imbalance of DE-OS[7:1]is the likeliest explanation.

The BiLSTM achieved its highest \enquote*{Hate} F1 on EN-OS[1:1], and its highest German \enquote*{Hate} F1 scores on DE-US[1:1] and DE-OS[1:1]. The two German datasets with imbalanced distributions yielded a slightly poorer performance in the \enquote*{Hate} class, similar to what was observed with the CNN. It is additionally worth noting that although the BiLSTM achieved similar \enquote*{Hate} F1 scores on DE-US[2:1] and DE-OS[7:1], its \enquote*{noHate} precision and recall on the latter dataset were lower than those from the former. This indicates that for DE-OS[7:1] the BiLSTM could only achieve good performance in the minority class by overfitting to it. Taken together with our observations from the CNN, this illustrates the detrimental effect of an imbalanced class ratio within small corpora.

mBERT had the best overall performance among the three architectures. Similar to the trend shown by the previous models, it achieved its highest macro-average F1 score on the EN-OS[1:1] dataset, but also its highest classwise scores as well. This benefit could have been due to the larger size of EN-OS[1:1] compared to the other corpora. The fact that scores for each class ratio also tended to be higher with the English datasets points to the model's strength with English training data, despite its multilinguality. Classwise performance on the [7:1] and [2:1] datasets is slightly stronger in the \enquote*{noHate} class than in \enquote*{Hate}, reflecting the datasets' skew towards \enquote*{noHate}. 

Among the [1:1] datasets, mBERT's classwise \enquote*{Hate} scores and macro-average F1 scores tended to be higher for the oversampled versions of a particular language than for the undersampled versions. For example, mBERT achieved a \enquote*{Hate} F1 of 63.7 on DE-OS[1:1] compared to 62.9 on DE-US[1:1]. The oversampled dataset also yielded better \enquote*{noHate} recall and F1, as well as better macro-average scores. The same pattern is observed with EN-US[1:1] and EN-OS[1:1], with the latter dataset giving significantly better scores in every category.

In addition, despite DE-OS[1:1] and EN-OS[1:1] having an identical class ratio, mBERT's much higher scores with the latter training set point to the necessity of a large amount of data for this architecture. However, the transformer's significantly higher classwise \enquote*{Hate} scores show that it is generally better able to cope with smaller dataset sizes than the BiLSTM and CNN. Among the three architectures examined, mBERT was the most successful at maintaining good minority-class performance on our relatively small corpora, making this architecture the better choice for low-resource setups. 

Although all three architectures achieved their best English \enquote*{Hate} F1 scores on the oversampled, balanced EN-OS[1:1], only mBERT had the same success in German with DE-OS[1:1]. The CNN's German \enquote*{Hate} F1 was the highest with DE-US[2:1], while the BiLSTM's was with DE-US[1:1]. This indicates that having a balanced class distribution is not the sole deciding factor for good minority-class performance, at least for small corpora. Among the [1:1] German datasets, the use of oversampling or undersampling did not play a deciding role for \enquote*{Hate} F1 performance. The difference between the \enquote*{Hate} F1 scores of EN-OS[1:1] and EN-US[1:1] was much higher, suggesting that oversampling the minority class might be a better option than undersampling the majority if the majority class is significantly larger. Additionally, our experiments indicate that the duplicated examples present in the oversampled datasets did not pose a significant problem for our models. More research will have to be done to confirm these conclusions, as well as to shed light on the exact interplay between class distribution and dataset size on minority class performance.

\begin{table}[t]
\centering
\resizebox{0.99\textwidth}{!}{
\begin{tabular}{ccccccccccc}
\toprule
\multicolumn{1}{c}{\textbf{Trainset}} & \multicolumn{1}{c}{\textbf{Accuracy}} & \multicolumn{3}{c}{\textbf{noHate}} & \multicolumn{3}{c}{\textbf{Hate}} & \multicolumn{3}{c}{\textbf{Macro-Avg}} \\ 
\midrule
&  & \textbf{P} & \textbf{R} & \textbf{F1} & \textbf{P} & \textbf{R} & \textbf{F1} & \textbf{P} & \textbf{R} & \textbf{F1} \\
DE-OS[7:1] &  74.48	&81.65&	86.96&	84.22&	38.48&	29.44&	33.36&	60.06&	58.2&	58.79
 \\
DE-US[2:1] & 71.25	&85.38	&76.36&	80.62&	38.21&	52.78	&44.33&	61.8&	64.57&	62.47

 \\
DE-US[1:1] &  69.19	&85.55	&72.98	&78.77&	36.26&	55.5&	43.86&	60.9&	64.24&	61.31

 \\
 DE-OS[1:1] &  77.6&	83.78&	88.54	&86.1	&47.95&	38.13&	42.48&	65.87&	63.33&	64.29
 \\
 \midrule
EN-TRAIN[7:1] &  58.49	&77.95	&66.00	&71.48	&19.47	&30.58	&23.79	&48.71	&48.29	&47.64
\\
EN-US[2:1] & 59.02&	77.41	&67.78	&72.27	&18.08	&26.45	&21.48	&47.75	&47.11	&46.88
 \\
EN-US[1:1] &  78.16	&97.06&	77.28&	86.05&	35.33&	84.13&	49.77&	66.2&	80.71&	67.91

 \\
EN-OS[1:1] &  87.35&	95.97&	89.23&	92.48&	50.54&	74.6&	60.26&	73.25&	81.92&	76.37
 \\
\botrule
\end{tabular}
}
\caption{Monolingual CNN performance after training on the various sampled datasets.}
\label{monocnn}
\end{table}

\begin{table}[t]
\centering
\resizebox{0.99\textwidth}{!}{
\begin{tabular}{ccccccccccc}
\toprule
\multicolumn{1}{c}{\textbf{Trainset}} & \multicolumn{1}{c}{\textbf{Accuracy}} & \multicolumn{3}{c}{\textbf{noHate}} & \multicolumn{3}{c}{\textbf{Hate}} & \multicolumn{3}{c}{\textbf{Macro-Avg}} \\ 
\midrule
&  & \textbf{P} & \textbf{R} & \textbf{F1} & \textbf{P} & \textbf{R} & \textbf{F1} & \textbf{P} & \textbf{R} & \textbf{F1} \\
DE-OS[7:1] &  24.97	&74.65	&8.26	&14.87	&20.19	&89.22	&32.93	&47.42	&48.74	&23.90
\\
DE-US[2:1] & 72.59&	81.21&	84.45&	82.8&	35.29&	30.27&	32.59&	58.25&	57.36&	57.7

 \\
DE-US[1:1] &  61.89	&81.87	&65.78	&72.95&	28.21&	47.99&	35.54&	55.04&	56.89&	54.24

 \\
DE-OS[1:1] & 72.71 & 81.81	&83.65	&82.72	&36.57&	33.64&	35.04&	59.19&	58.64&	58.88

 \\ 
 \midrule
 EN-TRAIN[7:1] & 67.96&	95.61&	66.28&	78.28&	25.77&	79.37&	38.91&	60.69&	72.82&	58.6

 \\

EN-US[2:1] & 81.02	&92.39&	85.25&	88.67&	34.38&	52.38&	41.51&	63.38&	68.81&	65.09

 \\
EN-US[1:1] & 63.88&	97.71&	59.95&	74.31&	25.0&	90.48&	39.18&	61.35&	75.21&	56.74

 \\
EN-OS[1:1] & 79.59	&93.37&	82.44&	87.56&	33.63&	60.32&	43.18&	63.5&	71.38&	65.37

 \\
\botrule
\end{tabular}
}
\caption{Monolingual BiLSTM performance after training on the various sampled datasets.}
\label{monolstm}
\end{table}

\begin{table}[t]
\centering
\resizebox{0.99\textwidth}{!}{
\begin{tabular}{ccccccccccc}
\toprule
\multicolumn{1}{c}{\textbf{Trainset}} & \multicolumn{1}{c}{\textbf{Accuracy}} & \multicolumn{3}{c}{\textbf{noHate}} & \multicolumn{3}{c}{\textbf{Hate}} & \multicolumn{3}{c}{\textbf{Macro-Avg}} \\ 
\midrule
&  & \textbf{P} & \textbf{R} & \textbf{F1} & \textbf{P} & \textbf{R} & \textbf{F1} & \textbf{P} & \textbf{R} & \textbf{F1} \\
DE-OS[7:1] & 75.00	&77.54	&87.42	&82.19	&67.62	&50.92	&58.09	&72.58	&69.17	&70.14 \\
DE-US[2:1] & 73.50	&81.14	&77.94	&79.51	&60.28	&64.89	&62.50	&70.71	&71.42	&71.00
 \\ 
DE-US[1:1] &  71.72	&82.61	&72.36	&77.14	&56.81	&70.47	&62.90	&69.71	&71.41	&70.02 \\ 
DE-OS[1:1] &  74.15	&81.93	&78.03	&79.93	&61.01	&66.64	&63.70	&71.47	&72.33	&71.81
 \\ 
\midrule 
 EN-TRAIN[7:1] &84.94	&86.11	&96.44	&90.99	&76.12	&42.15	&54.26	&81.12	&69.3	&72.62
   \\
EN-US[2:1] & 84.06	&92.84	&86.44	&89.53	&59.87	&75.21	&66.67	&76.35	&80.83	&78.10
\\
EN-US[1:1] & 78.11	&89.93	&81.33	&85.41	&48.78	&66.12	&56.14	&69.35	&73.72	&70.78 \\
EN-OS[1:1] &  99.12	&100.00	&98.89	&99.44	&96.03	&100.00	&97.98	&98.02	&99.44	&98.71 \\
\botrule
\end{tabular}
}
\caption{Monolingual mBERT performance after training on the various sampled datasets.}
\label{monobert}
\end{table}

\section{Conclusion}\label{conclusion}

Building automatic hate speech detection systems for low-resource languages is difficult due to the small amount of available datasets. Our goal in this paper was to investigate whether cross-lingual transfer learning could be used to mitigate the problem of data scarcity. We chose an English dataset with a broad hate speech definition for training and a similar German corpus for testing. Although the datasets were similar, we had to simplify the complex annotation schema of the target language dataset into the binary schema of the source dataset to make them compatible for the cross-lingual experiments. Our results showed that cross-lingual transfer learning is indeed an effective tool for hate speech detection in low-resource languages. Additionally, we assembled two corpora of previously-unseen, unlabeled target language data and applied an ensemble of trained classifiers to them. We showed that fine-tuning on these automatically-labeled examples improved classification performance, particularly within the hate speech class.
Additionally we investigated the issue of class imbalance in hate speech datasets. We produce several over- and undersampled datasets based on our English and German corpora, using class ratios that reflect the original datasets' ratios. We test the efficacy of oversampling compared to undersampling and conclude that both may possess advantages for specific dataset scenarios. 
Our goal for the future is to apply cross-lingual transfer learning to other language pairs with greater syntactic differences than German and English. In addition, since the differences of labeling schemas across various hate speech datasets could prevent the application of transfer learning methods, we aim to develop a method that can effectively combine datasets with different labeling schemas without the need for label modifications.

\bibliography{eacl2021}

\end{document}